\documentclass[runningheads]{llncs}
\usepackage{graphicx}
\usepackage{times}
\usepackage{latexsym}
\usepackage{url}
\usepackage{graphicx}
\usepackage{algorithm}
\usepackage{algorithmic}
\usepackage{booktabs}
\usepackage{amsfonts}
\usepackage{amsmath}
\usepackage{soul}
\usepackage{cite}

\begin{document}
\title{Imbalanced Sentiment Classification Enhanced with Discourse Marker}
%
%
\author{Tao Zhang\inst{1,2} \and
Xing Wu\inst{1,2} \and
Meng Lin\inst{1} \and
Jizhong Han\inst{1} \and
Songlin Hu\inst{1,2} }
\authorrunning{T. Zhang et al.}
%
\institute{Institute of Information Engineering, Chinese Academy of Sciences, Beijing, China \and
University of Chinese Academy of Sciences, Beijing, China \\
\email{\{zhangtao,wuxing,linmeng,hanjizhong,husonglin\}@iie.ac.cn}}
%
\maketitle              
\begin{abstract}
Imbalanced data commonly exists in real world, espacially in sentiment-related corpus, 
making it difficult to train a classifier to distinguish latent sentiment in text data.
We observe that humans often express transitional emotion between two adjacent discourses 
with discourse markers like "but", "though", "while", etc, and the head discourse and the tail discourse
\footnote{In this paper, we use the term "head discourse" to denote the sentence before the discourse marker and "tail discourse" to denote the sentence after the discourse marker.} 
usually indicate opposite emotional tendencies.
Based on this observation, 
we propose a novel plug-and-play method, which first samples discourses 
according to transitional discourse markers and then validates sentimental polarities 
with the help of a pre-trained attention-based model.
Our method increases sample diversity in the first place, can serve as a upstream preprocessing part in data augmentation.
We conduct experiments on three public sentiment datasets, 
with several frequently used algorithms.
Results show that our method is found to be consistently effective, even in highly imbalanced scenario,
and easily be integrated with oversampling method to boost the performance on imbalanced sentiment classification.


\end{abstract}
\section{Introduction}
Nowadays, people tend to express their feelings in online websites.
Such information often contains sentiment and opinions towards specific target such as products, services and events.
Analyzing these information can bring insights on what people need and can be  
helpful for both academic studies and industrial service. 
However, imbalanced data commonly exists in plenty of scenarios, making it hard to be utilized directly.

This phenomenon emerges frequently in sentiment-related area, 
where individuals tend to select and share content based on 
what the majority agrees with. 
In the review scenario, sentiment expressed by people is high imbalanced 
because they share similar evaluation standard on a relatively stable review object.
For example, the service provided by the same restaurant is unlikely to change too much
in a specific period, so the customer's reviews towards service 
are likely to be consistently good or bad except some extreme cases. 

Supervised machine learning techniques are widely used for sentiment classification 
and achieve better results than traditional lexicon-based methods.
Recently, deep learning models have shown promising results in this area. 
Nevertheless, the probability-based model is easily biased when the data distribution is highly imbalanced, 
which is often the case in sentiment classification task. 
As the imbalanced-ratio
\footnote{We define imbalanced-ratio to be:
number of samples in majority class $/$ number of samples in minority class.} increases, 
the performance of these models drops significantly~\cite{imb_2011_cikm}.


A natural solution to this problem is utilizing data augmentation to balance the dataset. 
Existing work can be divided into three categories.
Several studies on this matter have verified the effectiveness of re-sampling techniques, 
which either oversample data from the minority class~\cite{oversample_smote} or undersample data from the majority class~\cite{imb_2011_cikm}.
Generation-based methods employ deep generative models such as GAN or VAE 
to generate sentences from a continuous space with desired attributes of sentiment and tense.
Replacement-based methods generate sentences with different polarities by replacing sentimental words with synonyms and antonyms.
However, given a dataset which is highly imbalanced, all these methods are not good enough to achieve satisfying performance. 

We observe that humans often express transitional emotion between two adjacent discourses 
with discourse marker like "but", "though", "while", etc, and these two discourses 
usually indicate opposite emotional tendencies. 
Specifically, given a positive sentence "We got some lazy service, but the food is decisions.", 
the head discourse clearly shows negative emotion towards service and 
the tail discourse expresses positive emotion towards food, together they constitute a sentence more inclined to 
a positive expression.
This kind of transitional discourse marker generally emphasizes the meaning of the tail discourse.
In sentimental scenario, the expressed sentiment of the whole sentence is inclined to that of the tail discourse,
which can be used to generate discourse samples.

\begin{figure}[h]
	\begin{centering}
	\includegraphics[width=0.7\textwidth]{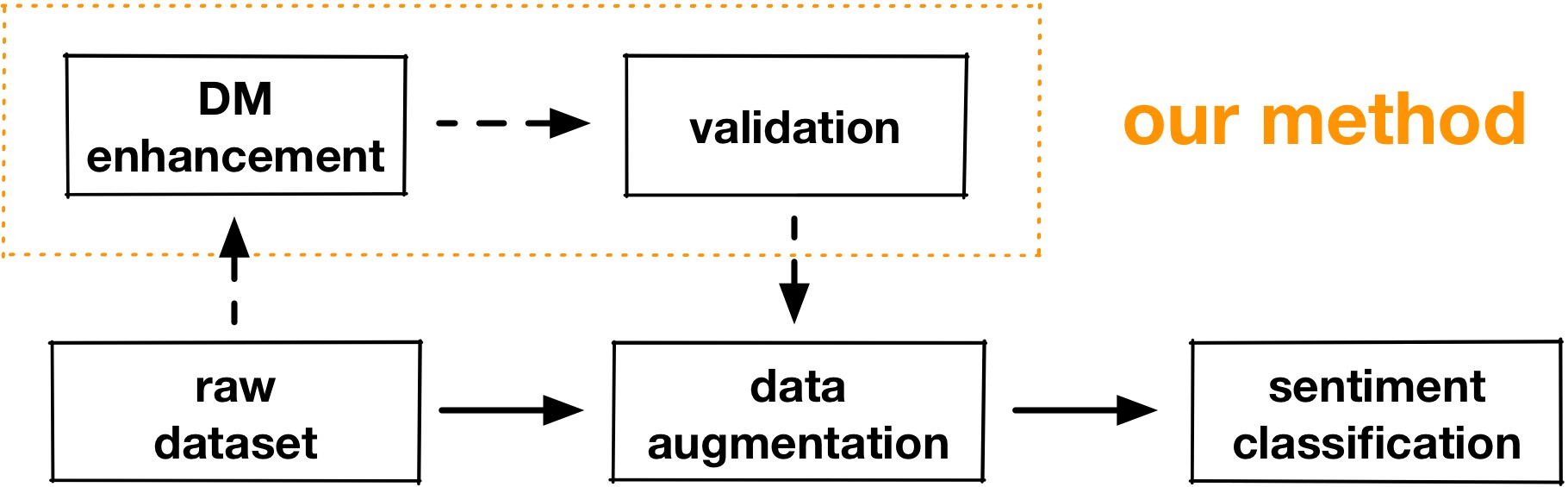}
    \caption{Simple illustration of our data augmentation with our method.
    The dashed arrows show that our method is plug-and-play and easily serves as a 
    upstream preprocessing part in data augmentation.
} \label{our method}
	\end{centering}
\end{figure}

Based on the above observation, we propose a novel plug-and-play method.
Firstly, we sample discourses connected by transitional discourse markers from original data. 
We annotate the head discourse with a opposite label and the tail discourse with original label from sampled sentence. 
Secondly, to prevent the problem of incorrect annotation, we introduce a pre-trained attention-based model to validate 
whether the semantics of generated samples are consistent with their label.
Finally, we obtain a expanded dataset with relatively low imbalanced-ratio and
we can further use other approaches like oversampling to rebalance the dataset.
The whole process is illustrated in Figure \ref{our method}.

Our contributions are listed as follows:
\begin{itemize}
    \item We propose a novel method that can be easily integrated with existing work to tackle the problem of imbalanced sentiment classification.
    \item To the best of our knowledge, this is the first work that apply discourse markers to data augmentation.
    \item We conduct experiments on several public datasets and results show that our method can boost the performance enormously.
\end{itemize}

\section{Related Work}

\subsection{Sentiment Classification}
Sentiment classification techniques can be roughly divided into three categories: 
lexicon-based approaches, traditional machine learning approaches and deep learning approaches.
As lexicon alone is insufficient to determine sentence-level or document-level sentiments, 
it is usually used as features in other methods, serving as a prior language knowledge.
Traditional machine learning approaches like support vector machine, naive bayes and decision tree 
extract words from text as features and model latent relationships between features to separate samples of different sentiments.  

Recently, models such as the Convolutional Neural Network (CNN)~\cite{deep-cnn-sentiment}, 
the Recurrent Neural Network (RNN)~\cite{rnn-sentiment}, memory network~\cite{memory-sentiment}
have been introduced for sentiment classification tasks.
These methods are endowed with powerful computation ability, 
capable of learning high-level features directly from low-level continuous representations. 
Apart from diverse models, language model pre-training has attracted wide attention and
fine-tuning on pre-trained language model has shown to be effective for improving many downstream natural language processing tasks.
BERT~\cite{BERT} obtained new state-of-the-art results on a broad range of diverse tasks by introducing deep masked language model.
Such pre-trained model can be used as sentiment classifier, 
transferring pre-learned knowledge to current problem.

\subsection{Data Augmentation}

Sampling-based methods are the most common ones in data augmentation. 
Undersampling~\cite{undersampling} randomly discards original samples from the majority class until the desired class distribution is obtained.
Undersampling leads to data waste and suffers from the problem of information loss.
On the contrary, oversampling repeatedly chooses samples from minority class,
resulting in overfitting because of lacking data diversity.
Some adaption of oversampling like SMOTE~\cite{oversample_smote} generates new samples by interpolating between  
pairs of the closest minority neighbors on continuous space. 

Some work solve this problem with generative models.
Hu et al.~\cite{TCGT} utilizes auto-encoder backbone to generate attribute controlled text.
Jia et al.~\cite{adversarial_examples} creates adversarial examples by adding distracting sentences to the input paragraph. 
Xie et al.~\cite{data_noising} considers noising primitives as a form of data augmentation for recurrent neural network-based language models. 
Generative methods can generate new samples. But limited by the model's capability, the quality of generated sentence can not be guaranteed.

Replacement-based methods can be seen as another paradigm, widely used in generation task.
The replacement process contains two steps, where to replace and how to replace.
Zhang et al.~\cite{char-cnn} extracts all replaceable words from the given text and randomly chose some of them to be replaced, 
then substitutes the replaceable words with synonyms from WordNet. 
Fadaee et al.~\cite{rare_word} focuses on the rare word problem in machine translation, replacing words in a source sentence with only rare words.
In unpaired sentiment-to-sentiment translation task, 
Xu et al.~\cite{unpaired} first deletes sentiment-related words and then uses a decoder to restore sentences based on attribute labels.

Imbalanced sentiment classification was first discussed in ~\cite{imb_2011_cikm}, where several sampling techniques are applied.
Their study demonstrates that imbalanced data distribution severely damages the performance of classifiers and 
sampling-based methods can significantly alleviate this situation.
Liu et al.~\cite{two-step} proposes that the challenge is how to obtain useful information of minority class from a imbalanced dataset.
Wu et al.~\cite{imb_2018_cikm} undersamples majority class to form several balanced subsets, then trains multiple classifiers which work together 
via an ensemble manner. In addition, they incorporate prior knowledge of sentiment expressions extracted from 
both existing sentiment lexicons and massive unlabeled data to enhance the learning of sentiment classifier.

\subsection{Discourse Marker}
Discourse marker was first proposed in~\cite{DisSent}, indicating deep conceptual relations between sentences.
Human writers naturally use a small set of very common words like "but", "and", to identify the relations between adjacent ideas.
They propose a discourse marker prediction (DMP) task, leveraging a high-level relationship between sentences to learn a meaningful representations.
The task aims to predict which discourse marker is used by the author to connect the two adjacent sentences.
Dozens of discourse markers are in discussion, including most common ones such as \textit{and, as, but, if, before} and less frequent ones.
We observe that transitional discourse markers can be view as a sentiment transition signal, 
and adjacent discourses usually indicate opposite emotional feelings.
Thus, the discourses connect by transitional discourse markers can be used to augment imbalanced datasets.

\begin{algorithm}[h]
    \caption{Detailed process of using our method for data augmentation.} \label{alg1}
    \begin{algorithmic}[1]
      \STATE Pretrain a sentiment classifier $\cal{C}$
      \STATE Given a imbalanced sentiment dataset $\cal{D}$ and a set of transitional discourse marker $M$
      \STATE Choose sentences from $D$ according to $M$, constituting ${\cal{D}}_m$
      \STATE Generate samples from ${\cal{D}}_m$ with \textbf{Swap} and \textbf{Crop} operations, constituting ${\cal{D}}_g$
      \FOR {each sample $s$ with label $l$ in ${\cal{D}}_g$}
        \STATE Use $\cal{C}$ to get predicted sentiment $l^{'}$
        \IF {$l$ is the same as $l^{'}$}
            \STATE Add $s$ to $\cal{D}$
        \ENDIF 
      \ENDFOR
      \STATE Then, use other data augmentation methods like oversampling to get a balanced dataset ${\cal{D}}_{b}$
      \STATE Perform sentiment classification on ${\cal{D}}_{b}$
    \end{algorithmic}
  \end{algorithm}

\section{Method}
In this section, we present our novel data augmentation method, most applicable to imbalanced sentiment classification.
Firstly, we give a detailed discussion about different discourse markers and 
explain why transitional discourse markers work in such scenario.
Then we introduce an attention-based method to validate generated samples, remedying for cases with false sentiment label.
After filtering out inconsistent samples, we combine generated samples with original dataset
to form a expanded dataset. Further operation like sampling-based method can be carried out, 
making it a balanced dataset for sentiment classification.
The details of our method are shown in algorithm \ref{alg1}.

\subsection{Discourse Marker Enhancement}
Words like \textit{and, but, when, if}, etc., serving as connection between two adjacent discourses, are called conjunction originally.
Nie et al.~\cite{DisSent} introduces the term "discourse marker" to name them,
indicating deep conceptual relations between discourses. 
A discourse can be one or several sentences, forming a relatively complete semantic sentence group.
Different discourse markers serves as different roles in sentences.
Based on their functions, we can divide them into fine-grained categories.

\begin{itemize}
    \item Time: when, before, after, then
    \item Parallel: and, also
    \item Transition: but, though, although, still
    \item Other functions: ...
\end{itemize}





We can observe that some discourse markers show strong indication of semantic relationship between adjacent discourses, such as \textit{and, but, because},
while some function in syntax aspect, such as \textit{when, before}. 
Further observation leads to a interesting phenomenon that discourses linked by transitional discourse markers often indicate
opposite sentiment tendencies when expressing emotion.
For example, in binary sentiment classification scenario, 
given a sentence "The actress is beautiful, but the plot is terrible" with negative label,
we can derive three sentences with clear sentiment inclination:

\begin{itemize}
    \item Swap the discourses: "The plot is terrible, but the actress is beautiful" emphasizes the tail discourse,
     making it a positive sample.
    \item Crop the head discourse: "The actress is beautiful" is clearly a positive sample, contrary to original sentence.
    \item Crop the tail discourse: "The plot is terrible" is no doubt a negative sample, consistent with original sentence.
\end{itemize}

Based on this phenomenon, we can generate samples in both category and augment imbalanced datasets. 
Formally, given a compound sentence $<s_h, m, s_t>$ with label $l$, 
where $s_h, s_t$ denote the head discourse and the tail discourse
and $m$ denotes a transitional discourse marker, we want to generate new sample $s_h$ with label $\overline{l}$, 
$s_t$ with label $l$ and $<s_t, m, s_h>$ with label $\overline{l}$. $l$ and $\overline{l}$ denote opposite sentiment.

One self-evident principle about data augmentation is that the new sample must agree with its label.
For sampling-based methods, the principle is satisfied because we directly sample the data with its label.  
For our method, $s_h/s_t$ can't be new samples if they are sentiment-irrelevant. For example, given a sentence 
"we have arrived here for half an hour, still no waiter comes to serve." 
the head discourse states an objective fact and can't be a new sample alone. 
Thus, we need to validate whether discourse are sentiment-irrelevant or not before generate new samples from them.

\subsection{Attention-Based Validator}
In this section, we present an attention-based validator to judge whether generated samples are sentiment-irrelevant or not.
Our validator consists of two parts: a recurrent neural network (RNN) as backbone and an attention block.
RNN is a class of neural networks, capable of processing sequential information like text.
We use a bidirectional Long Short Term Memory network (LSTM)~\cite{LSTM}, an adaption of RNN,
to model text from both directions.

Formally, given a input sentence $S = (w_1,w_2,...w_T)$ of $T$ words with label $l$, 
words are first embedded into continuous vectors: $x = (x_1,x_2,...,x_T)$.
Then word embeddings are then fed into LSTM. Each cell in LSTM can be computed as follows:
\begin{align*}
    & f_t = \sigma(W_f x_t + U_f h_{t-1} + b_f) \tag{1} \\
    & i_t = \sigma(W_i x_t + U_i h_{t-1} + b_i) \tag{2}\\
    & o_t = \sigma(W_o x_t + U_o h_{t-1} + b_o) \tag{3}\\
    & c_t = f_t \odot c_{t-1} + i_t \odot tanh(W_c x_t + U_c h_{t-1} + b_c) \tag{4} \\
    & h_t = o_t \odot tanh(c_t) \tag{5}
\end{align*}
where all $W,U$ are projection matrices and $b$ is bias. 
$\sigma$ is the sigmoid function and $\odot$ denotes element-wise multiplication.
$h_t$ stands for the hidden output of LSTM at timestep $t$ corresponding to $x_t$.
$f, i, o$ are gates with different functions, spanning $[0,1]$.
For each word $w_i$, we obtain its hidden output by concatenating its forward state and backward state
of LSTM of both directions $h_i = [{\overrightarrow{h_i}^T};{\overleftarrow{h_i}^T}]^T $.

To better model long sentences and accurately catch their sentiments,
we utilize the attention mechanism, which is inspired by the fact that
human visual attention is able to focus on a certain region on an image with high resolution while perceiving 
the surrounding image in low resolution and then adjusting the focal point over time.
In natural language processing, the attention mechanism works in a similar way and helps the model to
learn which part of text should be paid more attention, resulting in more reasonable sentence representations.
Formally, after obtaining hidden outputs,
we align each hidden output $h_i$ to the context vector $v$ to get the attention weight:
\begin{equation}
    f(h_i, v) = tanh(W_h h_i + W_v v + b)       \tag{6}
\end{equation}
where $W_h,W_v,b$ are trainable parameters and $f$ is to calculate how much attention we should pay to $h_i$.
Then we calculate a normalized attention weights:
\begin{equation}
    \alpha_i = \frac{exp(f(h_i,v))}{\sum_jexp(f(h_j,v))}    \tag{7}
\end{equation}

The attention weights evaluates the contribution of each word to sentiment classification. 
Thus we use the weighted sum of hidden outputs $c$ to represent input sentence, and transform $c$ into
a probability distribution $y$ on sentiment class labels:
\begin{align*}
    & c = \sum_{i=1}^T \alpha_i h_i \tag{8} \\
    & y = softmax(W_s c + b_s)      \tag{9}
\end{align*}
where $W_s,b_s$ are also a projection parameter.
After well pre-trained, an attention-based model can be used to filter generated samples 
whose semantics is not consistent with the label.

\section{Experimental Settings}
In this section, we introduce the detailed settings of our experiments.
We present datasets in Section 4.1.
Baselines are introduced in Section 4.2.
The experiment details are shown in Section 4.3.

\subsection{Datasets}
We conduct experiments on three public sentimental review datasets.
In the last part of Section 3.1, we discuss the principle that data augmentation must agree with.
For our method, $s_h/s_t$ can't be new samples if they are sentiment-irrelevant.
Thus, we hope that discourses in a transitional compound sentence are more likely to be sentiment-relevant in the first place.

The review datasets are able to fulfill our expectation. 
When people make comments, the review content is often about feelings towards specific targets.
Moreover, in a transitional compound sentence, discourses in opposite emotions tend to 
express feelings about different aspects of review object, 
which makes the discourse alone more likely to be sentiment-relevant. 
Specifically, we choose three binary classification datasets:
\begin{itemize}
    \item \textbf{MR} Movie Review Data (MR) proposed by Pang and Lee~\cite{MR}, 
    consisting of 5,331 positive and 5,331 negative reviews. The reviews are generally in one sentence.
    We randomly choose 80\% of them as training set and remain 20\% as testing set. \\
    \item \textbf{SST2} Stanford Sentiment Treebank (binary version), 
    an extension of the MR but with train/dev/test splits. T
    he split is 6,920 samples for training, 872 for validation, and 1,821 for testing~\cite{SST2}.  \\
    \item \textbf{CR} Customer reviews of various products with 2406 positive and 1367 negative samples~\cite{CR}.
    We split the dataset following the same way as what we have done to \textbf{MR}.
\end{itemize}

Without loss of generality, we assume negative samples to be minority class
and randomly sample sentences in negative class to construct imbalanced datasets.


\subsection{Baselines}
For one specific algorithm, we apply it to each dataset in two different settings 
to prove that our method can significantly boost performance when integrated with oversampling.
Specifically:
\begin{itemize}
    \item balanced dataset with oversampling.
    \item balanced dataset with our method and oversampling.
\end{itemize}

For algorithms, we consider three traditional machine learning methods: support vector machine (SVM),
naive bayes (NB), logistic regression (LR) and two basic deep neural networks: 
convolutional neural network (CNN), recurrent neural network (RNN).

\subsection{Experiment Details}
All traditional machine learning models, i.e. SVM, NB, LR are implemented with sklearn~\cite{sklearn} in default settings.
For CNN model, we use filter size $[3,4,5]$, each with 100 filters. 
We use pre-trained word embeddings trained on Google News\footnote{https://github.com/mmihaltz/word2vec-GoogleNews-vectors} and make them trainable. 
The dropout layer with dropout rate $=0.5$ is appended to the convolutional module.
For RNN model, we use a single layer bidirectional LSTM with 256 hidden units.
Our attention-based validator is pre-trained on Yelp dataset~\cite{yelp}, with LSTM hidden unit $=32$ and Adam~\cite{adam} optimizer.
The transitional discourse markers we use are $DM=[but,although, though, however, yet]$. 
Our evaluation metric is binary classification accuracy.

\section{Experimental Results}

\subsection{Overall Effectiveness of Our method}
Following the experiment settings introduced in Section 4.2, we test the performance of listed methods.
with imbalanced-ratio $\lambda = 5$.
The results are shown in Table \ref{overall effectiveness}.
Our method generates more samples and builds a more balanced dataset for further oversampling,
consistently increasing the classification performance.
It worthy to note that SVM performs unstable, even unchanged in CR dataset.
We ascribe this to the special modeling process of SVM, which is not the focus of this paper. 
Since our propose is to verify the performance of our data augmentation method, 
we remove SVM in following experiments for comparison convenience.

\begin{table}[htbp]
    \small
    \centering
    \caption{Experimental results of several methods in different settings.
    The evaluation metric is accuracy (\%).
    ``raw'' denotes raw imbalanced sentiment datasets.
    ``w/'' represents ``with''.
    ``os'' denotes ``oversampling''.
    ``our'' denotes ``our method''.
    }\label{overall effectiveness}
    \begin{tabular}{p{1.5cm} p{2cm}  p{2cm} p{2cm} p{2cm} p{2.5cm}}
    \toprule
    Method      & Setting                   & MR    & SST2      & CR    & Avg improvement\\
    \midrule
    NB            & w/os                      & 72.79 & 73.31     & 74.19 & - \\
                & w/our + os                & 71.90 & 76.99     & 76.20 & 1.60\\
    \midrule
    LR            & w/os                      & 68.88 & 69.02     & 74.60 & -\\
                & w/our + os                & 67.95 & 71.14     & 76.20 & 0.93\\
    \midrule
    SVM            & w/os                      & 66.34 & 49.91     & 50.00 & -\\
                & w/our + os                & 50.00 & 69.41     & 50.00 & 1.05\\
    \midrule
    CNN            & w/os                      & 71.33 & 75.28     & 77.82 & -\\
                & w/our + os                & 74.14 & 79.95     & 81.45 & 3.70\\
    \midrule
    RNN            & w/os                      & 71.80 & 74.30     & 75.60 & -\\
                & w/our + os                & 75.34 & 79.39     & 77.02 & 3.35\\
    \bottomrule
    \end{tabular}
\end{table}

\subsection{Effectiveness of Validator}
To verify the effectiveness of the attention-based validator, we conduct a group of 
ablation test in following settings:
\begin{itemize}
    \item balanced dataset after discourse marker enhancement and oversampling, without attention-based validator.
\end{itemize}

Due to space limitations, we take CNN as a example and the details are shown in Table \ref{validator effectiveness}.
We can observe that validator indeed contributes a lot to the total performance. 
As discussed in the last part of Section 3.1, the generated sample must agree with its label. 
In our method, the discourse sample may state a simple fact and be sentiment-irrelevant.
Our attention-based model serves as a validator and filters out false labeled cases,
significantly improving the quality of expanded dataset.

\begin{table}[htbp]
    \small
    \centering
    \caption{Ablation test on attention-based validator. 
    ``wo/'' represents ``without''.
    ``val'' denotes ``validator''.
    ``full'' denotes our full model.
    }\label{validator effectiveness}
    \begin{tabular}{p{1cm} p{1cm} p{2cm}  p{2cm} p{2cm} p{2cm} p{2cm}}
    \toprule
    Method      & IR   &  Setting              & MR    & SST2      & CR    & Avg improvement\\
            \midrule
   CNN  &   5           & wo/val               & 72.84 & 74.84     & 79.43 & -\\
        &                & full                & 74.14 & 79.96     & 81.45 & 2.81\\
        &    20         & w/os                  & 66.18 & 70.12     & 62.70 & -\\
        &                & full                 & 70.95 & 74.74     & 68.75 & 5.15\\
        &    100         & w/os                 & 61.65 & 65.01     & 60.28 & -\\
        &                & full                 & 67.95 & 69.08     & 61.69 & 3.93\\
    \bottomrule
    \end{tabular}
\end{table}

\begin{table}[t!]
    \small
    \centering
    \caption{Experimental results on highly imbalanced datasets.
    ``IR'' denotes imbalanced-ratio.
    }\label{highly imbalanced comparison}
    \begin{tabular}{p{1cm} p{1.2cm} p{1.8cm}  p{1.5cm} p{1.5cm} p{1.5cm} p{2.5cm}}
    \toprule
    IR  &   Method      & Setting                   & MR    & SST2      & CR    & Avg improvement\\
            \midrule
    10  &   NB           & w/os                      & 68.15 & 68.86     & 69.75 & -\\
        &                & w/our + os                & 69.41 & 72.87     & 71.98 & 2.50\\
        &   LR           & w/os                      & 60.87 & 60.07     & 63.91 & -\\
        &                & w/our + os                & 63.74 & 65.24     & 70.96 & 5.33\\
        &    CNN         & w/os                      & 72.94 & 76.49     & 70.76 & -\\
        &                & w/our + os                & 74.51 & 79.74     & 75.40 & 3.15\\
        &    RNN         & w/os                      & 71.96 & 69.14     & 73.59 & -\\
        &                & w/our + os                & 71.90 & 77.05     & 76.01 & 3.42\\
        \midrule
    20  &   NB             & w/os                    & 61.44 & 61.55     & 63.30 & -\\
        &                & w/our + os                & 63.21 & 67.55     & 68.75 & 4.41\\
        &   LR             & w/os                    & 53.85 & 54.31     & 54.43 & -\\
        &                & w/our + os                & 59.72 & 60.13     & 63.31 & 6.86\\
        &    CNN            & w/os                   & 68.26 & 67.87     & 53.42 & -\\
        &                & w/our + os                & 70.92 & 74.74     & 68.75 & 8.23\\
        &    RNN            & w/os                   & 60.93 & 70.84     & 60.69 & -\\
        &               & w/our + os                 & 70.03 & 76.44     & 73.99 & 9.33\\
        \midrule
    50  &   NB             & w/os                    & 55.10 & 56.12     & 55.84 & -\\
        &                & w/our + os                & 61.76 & 65.35     & 64.52 & 8.19\\
        &   LR             & w/os                    & 50.52 & 50.74     & 51.41 & -\\
        &                & w/our + os                & 56.24 & 56.01     & 57.46 & 5.61\\
        &    CNN            & w/os                   & 62.27 & 54.09     & 51.20 & -\\
        &                & w/our + os                & 67.69 & 68.97     & 67.14 & 12.08\\
        &    RNN            & w/os                   & 52.29 & 53.10     & 52.82 & -\\
        &               & w/our + os                 & 67.48 & 71.06     & 63.91 & 14.75\\
        \midrule
    100 &   NB             & w/os                    & 51.61 & 51.78     & 52.01 & -\\
        &                & w/our + os                & 61.76 & 65.34     & 64.51 & 12.07\\
        &   LR             & w/os                    & 50.15 & 49.97     & 50.40 & -\\
        &                & w/our + os                & 56.24 & 56.01     & 57.46 & 6.40\\
        &    CNN            & w/os                   & 53.74 & 50.74     & 50.60 & -\\
        &                & w/our + os                & 67.95 & 69.08     & 61.69 & 14.55\\
        &    RNN            & w/os                   & 50.62 & 50.85     & 50.60 & -\\
        &               & w/our + os                 & 68.57 & 67.01     & 65.12 & 16.21\\
    \bottomrule
    \end{tabular}
\end{table}

\subsection{Highly imbalanced datasets}
In this section, we test our method in highly imbalanced scenario, compared with only using oversampling method.
We set the imbalanced-ratio $\lambda$ as different values. 
The details can be seen in Table \ref{highly imbalanced comparison}.
Significant improvements are consistently obtained by integrating our method with oversampling.
In extremely imbalanced scenario, i.e. $\lambda=50,100$, integrated method even shows more boost of performance.
From the table illustration we can also observe that when dataset gets highly imbalanced, 
the oversample method does not help the classification\footnote{In binary classification scenario, 50\% accuracy means random guess.}.

Oversampling method is limited to the sample quality of minority class.
When dataset gets extremely imbalanced, the problem of sample diversity deficiency becomes severe,
which damages the model performance enormously.
In comparison, our method is able to generate minority samples from both majority class and minority class.
As long as the sample sentences contain transitional discourse markers, they can be used to generate 
one sample with the same label and two samples with opposite label.
It is worthy to note that transitional discourse markers is almost equally appearing
in both positive and negative emotions,
which means we can stably get a certain amount of minority samples through our method.
In other words, the more skewed the data distribution is, the more minority samples we can obtain from transition sentences of majority class.

\section{Discussion and Future Work}

So far we have present how to utilize the transition relation between two adjacent discourses to augment datasets.
Sentiment is just one aspect of sentence semantics and
different kinds of discourse markers serve as different semantic indicator, 
which could certainly assist the research on a vast amount of NLP tasks, 
such as semantics compression and language inference. 
In the future, we plan to explore the functions of other discourse markers in other area.
Additionly, we will conduct more complete experiments on 
integrating our method with other data augmentation approaches and give a detailed analysis.

\section{Conclusions}
In this paper, we focus on data augmentation technique in imbalanced sentiment classification.
We propose a novel two-step method, 
which first generates new samples according to transitional discourse markers
and then validates polarity correctness with a pre-trained attention-based model.
The experimental results proves that the semantics conveyed by transitional discourse marker can 
be utilized to generate sentimental discourses.
Our method is simple and plug-and-play, serving as a upstream part in data augmentation.
Based on a expanded diverse dataset with a relatively low imbalanced-ratio, 
any other data augmentation methods can then rebalance it for further sentiment classification.

\bibliographystyle{splncs04}
\bibliography{mybib}

\end{document}